\documentclass{article}

\usepackage[final]{neurips_2024}

\usepackage[utf8]{inputenc} % allow utf-8 input
\usepackage[T1]{fontenc}    % use 8-bit T1 fonts
\usepackage{hyperref}       % hyperlinks
\usepackage{url}            % simple URL typesetting
\usepackage{booktabs}       % professional-quality tables
\usepackage{amsfonts}       % blackboard math symbols
\usepackage{nicefrac}       % compact symbols for 1/2, etc.
\usepackage{microtype}      % microtypography
\usepackage{xcolor}         % colors
\usepackage{siunitx}
\usepackage{enumitem}
\usepackage{multirow}
\usepackage{graphicx}
\usepackage{subfigure}

\title{Cancer-Net SCa-Synth: An Open Access Synthetically Generated 2D Skin Lesion Dataset for Skin Cancer Classification}

\author{%
Chi-en Amy Tai$^{1*}$ \quad Oustan Ding$^{1*}$ \quad Alexander Wong$^1$\\
$^1$University of Waterloo\\
\texttt{\{amy.tai, oustan.ding, alexander.wong\}@uwaterloo.ca}
}

\begin{document}

\maketitle

\begin{abstract}
In the United States, skin cancer ranks as the most commonly diagnosed cancer, presenting a significant public health issue due to its high rates of occurrence and the risk of serious complications if not caught early. Recent advancements in dataset curation and deep learning have shown promise in quick and accurate detection of skin cancer. However, current open-source datasets have significant class imbalances which impedes the effectiveness of these deep learning models. In healthcare, generative artificial intelligence (AI) models have been employed to create synthetic data, addressing data imbalance in datasets by augmenting underrepresented classes and enhancing the overall quality and performance of machine learning models. In this paper, we build on top of previous work by leveraging new advancements in generative AI, notably Stable Diffusion and DreamBooth. We introduce Cancer-Net SCa-Synth, an open access synthetically generated 2D skin lesion dataset for skin cancer classification. Further analysis on the data effectiveness by comparing the ISIC 2020 test set performance for training with and without these synthetic images for a simple model highlights the benefits of leveraging synthetic data to improve performance. Cancer-Net SCa-Synth is publicly available at \url{https://github.com/catai9/Cancer-Net-SCa-Synth} as part of a global open-source initiative for accelerating machine learning for cancer care.
\end{abstract}

\section{Introduction}
\label{sec:intro}
In the United States, skin cancer ranks as the most commonly diagnosed cancer, presenting a significant public health issue due to its high rates of occurrence and the risk of serious complications if not caught early~\cite{gruber2017skin}. Advancements in dataset curation and associated classification challenges have shown promise in quick and accurate detection of skin cancer~\cite{rotemberg2021patient}. Methods leveraging deep learning have recently grown in popularity and were shown to enhance performance of skin cancer detection~\cite{adegun2021deep, gouda2022detection, leeCancerNetSCaTailoredDeep2020}. However, the effectiveness of these deep learning models largely depends on the quality of their training datasets. As seen in Table~\ref{tab:data-results}, current open-source datasets have significant class imbalances which impedes the effectiveness of these deep learning models~\cite{cassidy2022analysis,wen2024data}. 

\begin{table*}[htbp]
 \caption{Overview of existing skin cancer datasets and their class distribution in the training set with values from HAM10000 obtained from~\cite{tschandl2018ham10000} and ISIC values obtained from~\cite{cassidy2022analysis}.}
 \label{tab:data-results}
 \centering
    \begin{tabular}{ c|c|c|c } 
    \hline
    \textbf{Dataset} & \textbf{Total Images} & \textbf{Total Benign} & \textbf{Total Melanoma} \\
    \hline
    ISIC 2017 & 2,000 & 1,626 & 374 \\
    ISIC 2018 & 10,015 & 8,902 & 1,113 \\
    ISIC 2019 & 25,331 & 20,809 & 4,522 \\
    ISIC 2020 & 33,126 & 32,542 & 584 \\
    HAM10000 & 10,015 & 8,902 & 1,113 \\
    Cancer-Net SCa-Synth & 10,000 & 5,000 & 5,000 \\
     \hline
\end{tabular}
\end{table*}

In healthcare, generative artificial intelligence (AI) models have been employed to create synthetic data to address patient privacy concerns and data imbalance in datasets by augmenting underrepresented classes and enhancing the overall quality and performance of machine learning models~\cite{murtaza2023synthetic,rodriguez2022synthetic}. For skin cancer, synthetic data was shown to help in skin cancer classification with a study conducted using semantic and instance masks to augment the lack of annotated data~\cite{bissoto2018skin}. Other research in this cancer domain have also shown that using synthetic data from generative adversarial networks (GANs) can help in skin cancer classification models, but they unfortunately do not publicize their synthetically generated image dataset that they used for experiments nor the code to generate these images~\cite{sedigh2019generating,ghorbani2020dermgan,la2022deep,beynek2021synthetic,kaur2021synthetic}.  

Even though GANs have gained immense popularity and shown incredible promise, they are computationally intensive to train and exhibit limited diversity in their outputs~\cite{rombach2022high}. To overcome these challenges, Stable Diffusion was developed~\cite{rombach2022high}. While extremely powerful, Stable Diffusion on its own lacked fine-tuning for personalized content and required a large dataset for generating highly specific images. As such, DreamBooth is a trainer that considers text as another input and paired with Stable Diffusion, fine-tunes the text-to-image diffusion model~\cite{ruiz2023dreambooth}.

Despite the fact that these recent improvements on the GAN architecture have shown success in a multitude of industries for generating synthetic data, there have been no current open-source studies that analyze the effectiveness of Stable Diffusion with DreamBooth for generating synthetic skin lesion images for skin cancer classification. In this paper, we build on top of previous work by leveraging these new advancements in generative AI, notably Stable Diffusion and DreamBooth. We introduce Cancer-Net SCa-Synth, an open access synthetically generated 2D skin lesion dataset for skin cancer classification. We provide further analysis on the data effectiveness by comparing the test set performance for training with and without these synthetic images for a simple model. Cancer-Net SCa-Synth is publicly available at \url{https://github.com/catai9/Cancer-Net-SCa-Synth} as part of a global open-source initiative for accelerating machine learning for cancer care.

\section{Methodology}
\label{sec:method}
\begin{figure*}
  \centering
  \includegraphics[width=\linewidth]{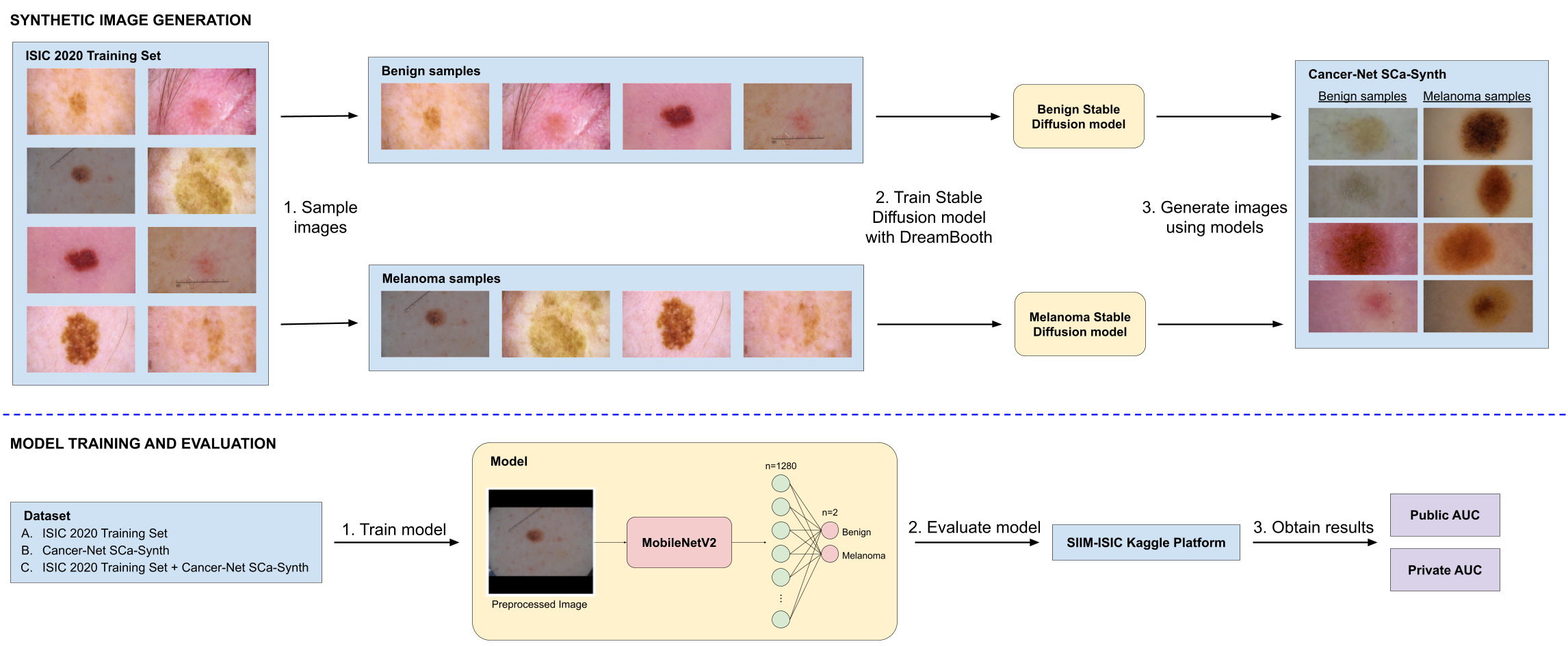}
  \caption{Process map of the entire model training pipeline from image generation to model evaluation.}
  \label{fig:process-map}
\end{figure*}

\subsection{Synthetic Image Generation}
As shown in the top portion of Figure~\ref{fig:process-map}, image generation was conducted using the Stable Diffusion model with the DreamBooth trainer. A separate Stable Diffusion model was trained for each of the two skin cancer classes. 300 randomly sampled benign skin cancer images from the ISIC 2020 training set~\cite{rotemberg2021patient} were used to train a Stable Diffusion model with the DreamBooth trainer for the benign class and 300 randomly sampled melanoma skin cancer images from the ISIC 2020 training set were used to train the melanoma model. A single word was used as the prompt for training, namely ``benign" and ``melanoma" for the associated models.

The learning rate was set to 5e-6 with a constant learning rate scheduler and no warmup steps. The image resolution used for training was 512x512 with no batch training used and a gradient accumulation step of 1. An AdamW optimizer was used with MSE loss and 2 epochs were initially used with a max 400 training steps. HuggingFace instead of Tensorflow/Keras was used as the generated images were often overfit and not realistic with Tensorflow/Keras. Using the associated trained models, 10,000 images were generated (5,000 for benign and 5,000 for melanoma). 

\subsection{MobileNetV2 Model Training}
\begin{figure}
    \hfill
    \subfigure[Original Image]{\includegraphics[width=0.45\linewidth]{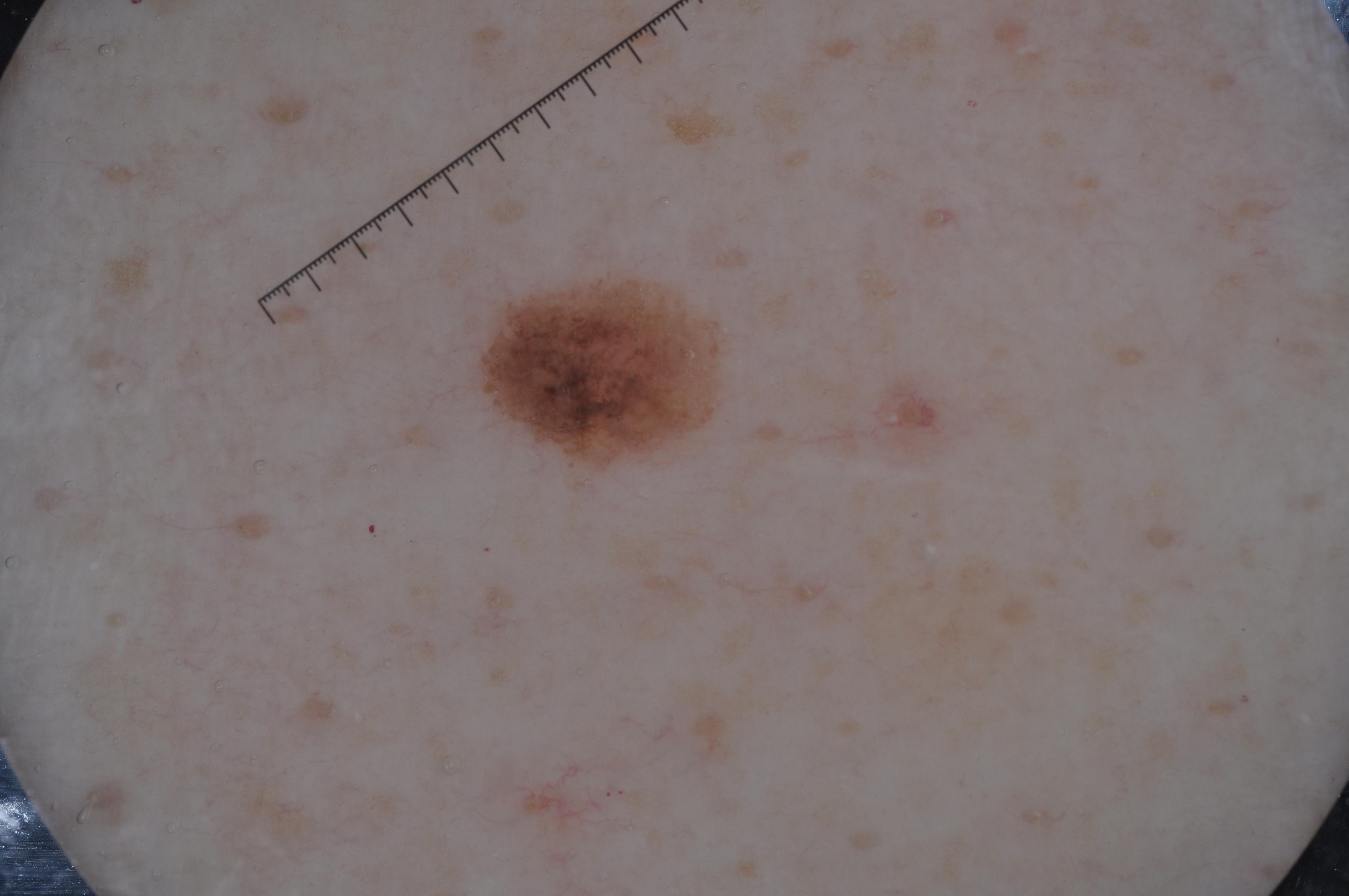}}
    \hfill
    \subfigure[Preprocessed Image]{\includegraphics[width=0.45\linewidth]{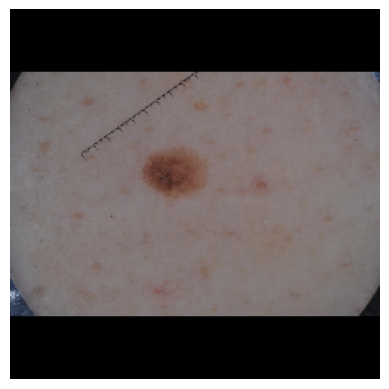}}
    \hfill
    \caption{Example of an image before (a) and after (b) preprocessing.}
    \label{fig:preprocess-image-example}
\end{figure}

The images were first standardized into a consistent format of size 224x224 with equal padding added to the height and/or width in instances with a smaller image. An example of an image before and after preprocessing is shown in Figure~\ref{fig:preprocess-image-example}.

We compare the performance using Cancer-Net SCa-Synth on the ISIC 2020 test set using a MobileNetV2 model for these scenarios: 

\begin{enumerate}[label=(\Alph*)]
    \item Train only on the ISIC 2020 training set
    \item Train only on Cancer-Net SCa-Synth
    \item Train with Cancer-Net SCa-Synth and fine-tune with ISIC 2020 training set
\end{enumerate}

For the first two scenarios, we used the ImageNet weights as the initial model weights. The pipeline was adapted from~\cite{kaggleTrainingMobileNet} with modifications for the specific dataset, binary classification, and class imbalance. We set the MobileNetV2 layers to be trainable and included a dropout layer of 0.3. We compute the area under the ROC curve (AUC) and accuracy and use a loss of binary cross-entropy with an AdamW optimizer. The initial learning rate was set to 0.0001 with an initial weight decay of 0.004. The data was split with the ratio 60:40 for the training:validation set and stratified with the skin cancer target label. 

The model was trained over 10 epochs with a batch size of 32 and the best model weights (based on the validation AUC) were used to evaluate the performance on the ISIC 2020 test set. Since the ISIC 2020 test set labels were not available online, we used the SIIM-ISIC Kaggle competition platform~\cite{pan_siim-isic_2020} to evaluate the performance on the test set. This platform provides two scores: a public score and a private score. The public score reflects the AUC performance on 30\% of the test data, while the private score measures the AUC performance on the remaining 70\% of the test data. The specific data split between the two subsets is unknown. The entire model training and evaluation flow can be seen in the bottom half of Figure~\ref{fig:process-map}.

\section{Results}
\label{sec:results}
\begin{table*}[htbp]
 \caption{The performance of various training data on the ISIC 2020 test dataset using a basic adapted MobileNetV2 architecture. In this table, ISIC 2020 refers to only the training set.}
 \label{tab:test-results}
 \centering
    \begin{tabular}{ c|c|c|c|c } 
    \hline
    \textbf{Scenario} & \textbf{Trained} & \textbf{Fine-Tuned} & \textbf{Private Score} & \textbf{Public Score} \\
    \hline
    (A) & ISIC 2020 & None & 0.6370 & 0.6475 \\
    (B) & Cancer-Net SCa-Synth & None & 0.5344 & 0.5194 \\
    (C) & Cancer-Net SCa-Synth & ISIC 2020 & 0.6776 & 0.7376 \\
     \hline
\end{tabular}
\end{table*}

As seen in Table~\ref{tab:test-results}, the test set performance was highest for both the private and public score in Scenario C. Notably, training using Cancer-Net SCa-Synth and fine-tuning with the ISIC 2020 training set outperforms training on either dataset alone by over 0.04 (private) and 0.09 (public). Sample images generated from the trained benign and melanoma Stable Diffusion models are shown in the top right of Figure~\ref{fig:process-map}. 

\section{Conclusion}
\label{sec:concl}
Given the rise of generative AI and its promising results in producing synthetic data for more effective model training, this paper builds on top of previous work by leveraging new advancements in generative AI, notably Stable Diffusion and DreamBooth. We introduce Cancer-Net SCa-Synth, an open access synthetically generated 2D skin lesion dataset containing 10,000 images for skin cancer classification that is equally distributed between the benign and melanoma skin cancer cases. Using a simple MobileNetV2 model, we demonstrate the benefits of leveraging synthetic data to improve performance and obtain an AUC that is higher than training on only the ISIC 2020 dataset by over 0.04 (private) and over 0.09 (public). 

\section{Future Work}
\begin{figure}
  \centering
  \includegraphics[width=\linewidth]{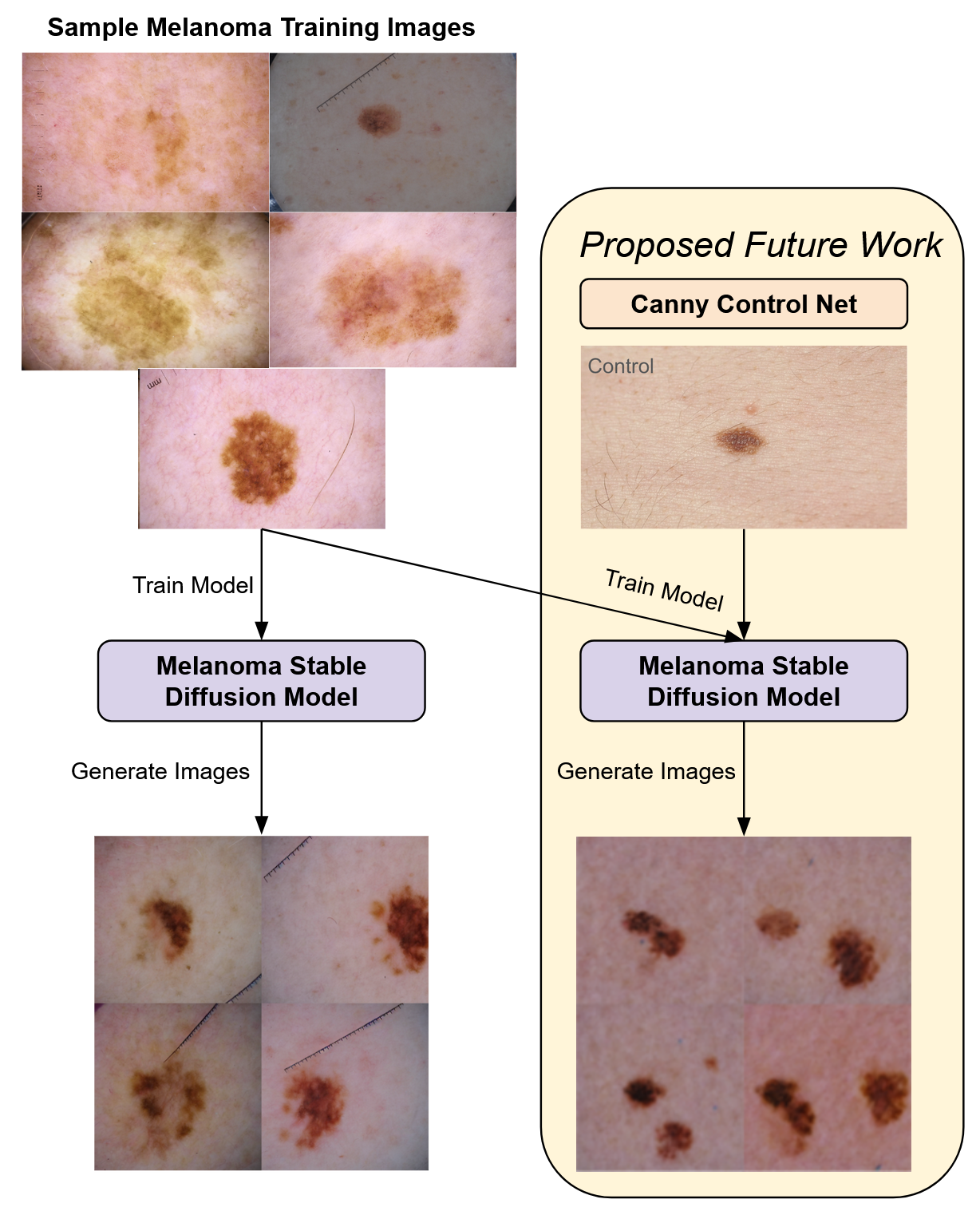}
  \caption{Comparison of the generated results with and without the use of ControlNet.}
  \label{fig:controlnet}
\end{figure}

Future work includes a comparison study with other deep learning models beyond MobileNetV2 and using ControlNet to add conditional control to the text-to-image diffusion models~\cite{zhang2023adding}. To illustrate the impact of training with ControlNet, a Stable Diffusion model was trained using DreamBooth for melanoma images, using five images to generate four sample images (Figure~\ref{fig:controlnet}, left). As seen in the right side of Figure~\ref{fig:controlnet}, by using a canny ControlNet to inpaint colors of a reference image, higher quality images can be generated.

{
\small

\bibliography{neurips_2024}
}

\end{document}